\documentclass{article}

\usepackage{microtype}
\usepackage{graphicx,dblfloatfix}
\usepackage{booktabs} 

\usepackage{caption}
\usepackage{subcaption}
\usepackage{paralist}
\usepackage{mdwlist}
\usepackage{hyperref}
\usepackage{mathtools}
\usepackage{amsmath,amsfonts}

\usepackage[accepted]{icml2019}

\icmltitlerunning{Non-Parametric Priors For GAN}

\begin{document}

\twocolumn[
\icmltitle{Non-Parametric Priors For Generative Adversarial Networks}



\icmlsetsymbol{equal}{*}

\begin{icmlauthorlist}
\icmlauthor{Rajhans Singh}{to}
\icmlauthor{Pavan Turaga}{to,ed}
\icmlauthor{Suren Jayasuriya}{to,ed}
\icmlauthor{Ravi Garg}{goo}
\icmlauthor{Martin W. Braun}{goo}

\end{icmlauthorlist}

\icmlaffiliation{to}{School of Electrical, Computer, and Energy Engineering, Arizona State University, Tempe, AZ, USA}
\icmlaffiliation{goo}{Intel Corporation, Chandler, AZ, USA}
\icmlaffiliation{ed}{School of Arts, Media and Engineering, Arizona State University, Tempe, AZ, USA}

\icmlcorrespondingauthor{Rajhans Singh}{rsingh70@asu.edu}
\icmlcorrespondingauthor{Pavan Turaga}{pturaga@asu.edu}
\icmlcorrespondingauthor{Suren Jayasuriya}{sjayasur@asu.edu}
\icmlcorrespondingauthor{Ravi Garg}{ravi.garg@intel.com}
\icmlcorrespondingauthor{Martin W. Braun}{martin.w.braun@intel.com}

\icmlkeywords{Machine Learning, ICML}

\vskip 0.3in
]



\printAffiliationsAndNotice{}  

\begin{abstract}
 The advent of generative adversarial networks (GAN) has enabled new capabilities in synthesis, interpolation, and data augmentation heretofore considered very challenging. However, one of the common assumptions in most GAN architectures is the assumption of simple parametric latent-space distributions. While easy to implement, a simple latent-space distribution can be problematic for uses such as interpolation. This is due to distributional mismatches when samples are interpolated in the latent space. We present a straightforward formalization of this problem; using basic results from probability theory and off-the-shelf-optimization tools, we develop ways to arrive at appropriate non-parametric priors. The obtained prior exhibits unusual qualitative properties in terms of its shape, and quantitative benefits in terms of lower divergence with its mid-point distribution. We demonstrate that our designed prior helps improve image generation along any Euclidean straight line during interpolation, both qualitatively and quantitatively, without any additional training or architectural modifications. The proposed formulation is quite flexible, paving the way to impose newer constraints on the latent-space statistics.
\end{abstract}

\section{Introduction}\label{introduction}


Advances in deep learning have resulted in state-of-the-art generative models for a wide variety of data generation tasks. Generative methods map sampled points in a low-dimensional latent space with known distribution to points in high-dimensional space with distributions matching real-data. In particular, generative adversarial networks (GANs)~\cite{goodfellow2014generative} have shown successful applications in super-resolution~\cite{ledig2017photo}, image-to-image translation~\cite{isola2017image, zhu2017unpaired}, text-to-image translation~\cite{reed2016generative}, image inpainting \cite{pathak2016context}, image manipulation~\cite{zhu2016generative}, synthetic data generation~\cite{shrivastava2017learning} and domain adaptation~\cite{tzeng2017adversarial}. 

A GAN architecture consists of a generator $G$ and a discriminator $D$. The generator $G$ maps low-dimensional latent points $z \sim \mathcal{P}_z$ to high-dimensional data distribution $\mathcal{P}_{data}$. The latent-space distribution $\mathcal{P}_z$ is typically chosen to be a normal or uniform distribution. The goal of the generator $G$ is to produce data such that they are perceptually indistinguishable from real data. However, the discriminator $D$ is trained to distinguish between `fake' and `real' data. Both the generator and discriminator are trained in an adversarial fashion, and at the end of training the generator learns to generate data with a distribution similar to the real one.

One natural question for generative models is how to model the latent space effectively to generate diverse and varied output. Interpolating between samples in the latent space can lead to semantic interpolation in image space~\cite{radford2015unsupervised}. Interpolation can help transfer certain semantic features of one image to another. Successful interpolation also shows that GANs do not simply over-fit or reproduce the training set, but generate novel output. Interpolation has been shown to disentangle factors of variation in the latent space with many applications \cite{liu2018exploring,kumar2018disentangling, fu2017learning, yin2017semi}.

Imposing a parametric structure on the latent space can cause distributional mismatches where the prior distribution does not match the interpolated point's distribution. This mismatch causes the interpolated points to lose fidelity in quality ~\cite{white2016sampling}. Previous research has resulted in various parametric models to fix this problem~\cite{white2016sampling,kilcher2017semantic,agustsson2017optimal}. One of the findings in prior work \cite{lesniak2018latent} is that the use of a Cauchy distributed prior solves the distributional mismatch problem. But, Cauchy is a very peculiar distribution, with undefined moments and a heavy-tail. This  means that during inference there will always be a number of undesirable outputs (as acknowledged also in \cite{lesniak2018latent}) due to latent vectors being sampled from the these tails.

In this paper, we propose the use of \textbf{non-parametric priors} to address the aforementioned issues. The advantage of a non-parametric prior is that we do not use any modeling assumptions and propose a general optimization approach to determine the prior for the task at hand. In particular, our contributions are as follows:
\begin{compactitem}
    \item We analyze the distribution mismatch problem in latent-space interpolation using basic probability tools, and derive a non-parametric approach to search for a  prior which can address the distribution mismatch problem.
    
    \item We present algorithms to solve for the prior using off-the-shelf optimizers, and show that obtained priors have interesting multi-lobe structures with fast decaying tails, resulting in mid-point distribution to be close to the prior.  

    \item We show that the resulting non-parametric prior yields better quality and diversity in generated output, with no additional training data nor any added architectural complexity. 
    
\end{compactitem}

More broadly, our approach is a general and flexible method to impose other constraints on latent-space statistics. Our goal is not to outperform all the latest developments in generative models, but to show that our proposed stand-alone formulation can boost performance with no added training or architectural modifications.

\section{Background and Related Work}
\textbf{Generative Adversarial Network:} As described in Section~\ref{introduction}, a GAN consists of two components: a generator $G$ and a discriminator $D$, which are adversarially trained against one another until the generator can map latent-space points to a high dimensional distribution which the discriminator cannot distinguish from true data samples.  Formally, this can be expressed as a min-max game in \eqref{eq:1} which the generator tries to minimize and the discriminator tries to maximize \cite{goodfellow2014generative}:
\begin{equation}
\label{eq:1}
\begin{aligned}
\min_{G} \max_{D} V(D,G) =  E_{x \sim \mathcal{P}_x}[\log(D(x))] \\
+ E_{z \sim \mathcal{P}_z}[\log(1-D(G(z)))],
\end{aligned}
\end{equation}

where, $V$ is the objective function, $x\sim \mathcal{P}_x$ are real data points sampled from a true distribution, and $z \sim \mathcal{P}_z$ are sampled points from the latent-space distribution. If the training of the GAN is stable and the Nash equilibrium is achieved, then the generator learns to generate samples similar to the true distribution. In general, GAN training is not always stable, thus several methods have been introduced to improve the training~\cite{salimans2016improved,arjovsky2017towards}. This includes different kinds of divergences and loss functions~\cite{nowozin2016f,arjovsky2017wasserstein, gulrajani2017improved}. Several other works improve generated image quality~\cite{dai2017towards, zhang2017stackgan} or resolution \cite{denton2015deep,karras2017progressive}.


\textbf{Interpolation:} For any two given latent-space points $z_1,z_2 \sim \mathcal{P}_z$, a linearly-interpolated point $z_\lambda$ is given by $z_\lambda = (1-\lambda)z_1 + \lambda z_2$ for some $ \lambda \in [0,1].$ It has been shown that GANs can generate novel outputs via linear interpolation, and as the line is traversed ($\lambda: 0 \to 1$), the output images smoothly transition from one to another without visual artifacts~\cite{radford2015unsupervised}. They further showed that vector arithmetic in the latent space has corresponding semantic meaning in the output space, e.g. latent-space points for ``man with glasses" - ``man without glasses" + ``woman without glasses" generates an image of a woman wearing glasses (c.f. Fig. 7 of~\cite{radford2015unsupervised}). 


\textbf{Distribution Mismatch of Interpolated Points:} Interpolation, while semantically meaningful, presents challenges in ensuring all interpolated points preserve the same data quality (or in the case of images visual quality). Most GANs utilize simple parametric distributions such as normal or uniform as the prior distribution to sample the latent space. However, these two choices of priors cause the interpolated point's distribution to not match with either the normal or uniform distribution as observed by~\cite{kilcher2017semantic}. We replicate this argument below for the sake of exposition, since this is the core problem we tackle in this paper.

Let $z_1, z_2 \sim \mathcal{N}(0,\sigma^2I)$ be two points in the latent space of the GAN's generator, and let $z_{\lambda} = (1-\lambda)z_1 + \lambda z_2$ be a linearly interpolated point. Note that $\sigma_{z_\lambda}^2 = (1-\lambda)^2\sigma_z^2 + \lambda^2 \sigma_z^2$. We are interested when $\sigma_{z_\lambda}^2 = \sigma_{z}^2$, which only holds for delta functions for finite moment distributions (we later prove this statement more formally in Section~\ref{approach}). However, we see that the worst case for when $\sigma_{z_\lambda}^2$ is different from $\sigma_z^2$ occurs at $\lambda = 1/2$ or the midpoint denoted $z^{,} = \frac{z_1 + z_2}{2}$. Analyzing the Euclidean norm of $z_1,z_2$ and $z^{,}$ gives the following equations~\cite{kilcher2017semantic}: 

\begin{equation}
\label{eq:3}
\begin{aligned}
z_1, z_2 \sim \mathcal{N}(0,\sigma^2 I) &\Rightarrow &
\Vert z_1 \Vert^{2},\Vert z_2 \Vert^{2} \sim \sigma^{2} \mathcal{X}^2(d),\\
z^{,}=\frac{z_1 + z_2}{2} &\Rightarrow& \Vert z^{,} \Vert^{2} \sim \frac{\sigma^{2}}{2} \mathcal{X}^2(d),
\end{aligned}
\end{equation}

where, $d$ is the dimension of the latent-space and $\mathcal{X}^2$ is the chi-squared distribution. The GAN is trained with latent-space whose norm squared distance follows a $\sigma^{2} \mathcal{X}^2(d)$ distribution according to \eqref{eq:3}. However, the mid-point will have a distribution $\frac{\sigma^{2}}{2} \mathcal{X}^2(d)$. Clearly, there is distribution mismatch between the points at which the GAN is trained to generate realistic samples and the mid-point where we want to do interpolation. Further, this distribution mismatch becomes worse if we increase the dimension of the latent space. Finally, it has been shown that the Normal distribution in higher dimensions forms a `soap bubble' structure, i.e. most of its probability mass concentrates in an annulus around the mean, rather than near the mean itself \cite{hall2005geometric}. This implies that interpolations that traverse near the origin of the latent space will suffer degradation in output fidelity/quality, which has been confirmed for GANs in~\cite{white2016sampling}. A similar proof for a distribution mismatch can be shown for the uniform distribution.

We note that we are not the first to propose a solution to this problem. Several prior approaches have proposed solutions, either through new interpolation schemes or new prior distributions different from normal or uniform that suffer less distribution mismatch. White et al. proposes spherical linear interpolation by following the geodesic curve on a hypersphere to avoid interpolating near the origin (and thereby minimize distribution mismatch) if the latent points are sampled uniformly on a sphere with finite radius \cite{white2016sampling}. However, this interpolation is not semantically meaningful if the path becomes too long and it passes through unnecessary images as noted in \cite{kilcher2017semantic}. Similar to spherical interpolation, \cite{agustsson2017optimal} propose a normalized interpolation. Yet, it inherits similar issues as in spherical interpolation in not being the shortest path.

Other approaches have attempted to define new prior distributions to ensure the interpolated points have low distribution mismatch. This is similar to the method we employ in our paper, except ours is non-parametric. Kilcher et al. propose a new prior distribution defined as follows:

\begin{equation}
    \label{eq:6}
    v \sim Uniform(S^{d-1}), r\sim \Gamma(\frac{1}{2},\theta), z = \sqrt{r}v
\end{equation}
 where $d$ is the dimension of the latent space, $\Gamma(\frac{1}{2},\theta)$ is the Gamma distribution and $z$ is a latent vector~\cite{kilcher2017semantic}. Latent spaces defined using this prior distribution do not suffer as much from mid-point distribution mismatch. Further work by Lesniak et al. showed that the Cauchy distribution $P(z) = \frac{1}{\pi(1+z^2)}$ induces a midpoint distribution which is the same as itself~\cite{lesniak2018latent}. However, the Cauchy distribution has undefined moments, which makes analysis difficult, and also is heavy-tailed, which can lead to undesirable outputs.

\section{Design of non-parametric priors for GANs}\label{approach}
The primary motivation to design non-parametric priors is that in order to have a prior whose distribution of mid-points is close to the original prior, we need to optimize for an appropriate cost over the space of density functions. This optimization is easily done for the non-parametric case, and requires rather few assumptions. Our terminology of {\em non-parametric} stems from classical density estimation approaches, rather than the more modern usage in Bayesian non-parametric.

Let $f_X(x)$ be the chosen prior distribution. Let $x_1$ and $x_2$ be two samples drawn from $f_X(x)$. Let an interpolated point be given by: $(1 - \lambda) x_1 +  \lambda x_2$, for $0 < \lambda < 1$. The precise relation between the distribution of this interpolated point and $f_X(x)$ is given analytically as follows.






\paragraph{Property 1:} If $x_1$ and $x_2$ are two independent samples drawn from $f_X(x)$, then the density function of $(1 - \lambda) x_1 +  \lambda x_2$, for $0 < \lambda < 1$ is given by $\frac{1}{|\lambda(1-\lambda)|}f_X(\frac{x}{\lambda}) * f_X(\frac{x}{1 - \lambda})$.

\paragraph{Proof:} The proof is a direct application of the following two results from probability theory.

\begin{compactitem}
    \item $R_1$: If $X_1$ and $X_2$ are two independent random variables, with density functions $f_{X_1}(x)$ and $f_{X_2}(x)$, then the density of their sum $X_1 + X_2$ is given by the linear convolution $f_{X_1}(x) * f_{X_2}(x)$.
    \item $R_2$: If random variable $X$ has density function $f_X(x)$, then for $\alpha \in \mathbb{R}$, the density of $\alpha X$ is given by $\frac{1}{| \alpha |} f_X(\frac{x}{\alpha})$.
\end{compactitem}

Apply $R_2$ to $
(1- \lambda) x_1$ and $\lambda x_2$ separately, then convolve the results using $R_1$. QED.

Following from here, the distribution mismatch problem can be expressed as the search for a prior distribution $f_X(x)$ such that the distribution of any other interpolated point is close to $f_X(x)$. That is, we would like $f_X(x)$ to satisfy:
\begin{align}
f_X(x) = \frac{1}{|\lambda(1-\lambda)|} f_X(\frac{x}{\lambda}) * f_X(\frac{x}{1 - \lambda}), \label{eq:constraint}
\end{align}

where, $\lambda \in (0,1)$. The only distributions we are aware of that satisfy this condition are the Cauchy (undefined moments, heavy-tailed), and delta functions (zero variance).

\paragraph{Property 2: } The only density functions with finite moments that satisfy condition \eqref{eq:constraint} are delta functions.

\paragraph{Proof:} A density function that satisfies condition \eqref{eq:constraint}, will also satisfy the equality of all moments of the left and right side densities. By specifically applying this to the equality of variances, we will show that the only solution is a delta function (among the class of finite moment densities). The following two results come handy.

\begin{compactitem}
    \item $R_3$: If $X_1$ and $X_2$ are two independent random variables, with respective variance $\sigma_1^2$ and $\sigma_2^2$, then the variance of their sum $X_1 + X_2$ is given by  $\sigma_1^2 + \sigma_2^2$.
    \item $R_4$: If random variable $X$ has variance $\sigma^2$, then for $\alpha \in \mathbb{R}$, the variance of $\alpha X$ is given by $\alpha^2 \sigma^2$.
\end{compactitem}
    
If \eqref{eq:constraint} holds, it must imply the equality of the variance of the prior, and the variance of any intermediate-point. i.e. $\sigma_X^2 = (1 - \lambda)^2 \sigma_X^2 + \lambda^2 \sigma_X^2$. If $\sigma_X$ is finite,  equality happens if and only if $\sigma_X = 0$. This implies that $f_X(x)$ is a delta function. QED.

The search for a density function that satisfies \eqref{eq:constraint} is thus not meaningful in the context of generative models, because a delta function as prior implies constant output. Cauchy, on the other hand, is too specific a choice, and suffers from pathologies such as undefined moments, which renders imposing any additional constraints on latent-space statistics impossible. It also is heavy-tailed, which may cause generation of undesirable outputs for samples from the tails. 

 What if we relax condition \eqref{eq:constraint}, such that we do not seek exact equality, but closeness of the left and right sides? The next section shows that this relaxed search results in a problem which can be solved using standard off-the-shelf function minimizers. Using this approach we obtain distributions with many useful properties. If we let $P(x) = f_X(x)$, and $Q(x;\lambda) = \frac{1}{\lambda(1-\lambda)} f_X(\frac{x}{\lambda}) * f_X(\frac{x}{1 - \lambda})$, we seek to minimize some form of distance or divergence between $P(x)$ and $Q(x)$ among densities with finite-variance. This optimization problem is defined in the next section. 

 
 \subsection{Searching for the optimal prior distribution}
As mentioned earlier, instead of enforcing exact equality as in condition \eqref{eq:constraint}, we would like to minimize the discrepancy between the left and right sides. A natural choice would be to minimize the KL divergence between $P(x) = f_X(x)$, and $Q(x;\lambda) = \frac{1}{\lambda(1-\lambda)} f_X(\frac{x}{\lambda}) * f_X(\frac{x}{1 - \lambda})$. Ideally, it might make sense to minimize this over the entire range of $\lambda$'s, i.e. 

\begin{align}
& \underset{P}{\text{minimize}} \int f(P(x) \| Q(x;\lambda)) d\lambda,
\end{align}

where $f$ is the chosen divergence/distance between the densities. However, this is likely an intractable problem due to integration over $\lambda$. In order to make this tractable, we observe that for a given $\lambda$, the mean of $Q(x;\lambda)$ is the same as the mean of $P(x)$. However, the variance of $Q(x;\lambda)$ goes as $(1 - \lambda)^2 \sigma_X^2 + \lambda^2 \sigma_X^2$. Thus, for interpolation problems, where $\lambda \in (0,1)$, the largest discrepancy in variance between $P$ and $Q$ occurs at a value of $\lambda = 0.5$. Thus, for interpolation problems, we suggest that minimizing for the worst-case error is sufficient. 

\subsection{Optimization problem for interpolation priors}
For interpolation priors, we minimize the KL divergence between $P(x)$ and $Q(x;\lambda = 0.5)$. To create a tractable problem, we restrict $P(x)$ to be defined over a compact domain, without loss of generality, we choose it to be $[0,1]$. We discretize the domain with sufficient fineness, typically we choose $2^{10}$ bins in $[0,1]$. The distribution is now discretely represented by the bin-centers $P = \{p_i\}_{i=1}^n$. The optimization problem now becomes:

\begin{align}
& \underset{P}{\text{min}}
 f(P \| Q) \text{ s. t.  } \sum_{i=1}^{n}p_i=1, \text{ and } p_i \geq 0.     \label{eq:8}
\end{align}

\begin{figure}[!htb]
     \centering
         \includegraphics[ width=0.49\textwidth]{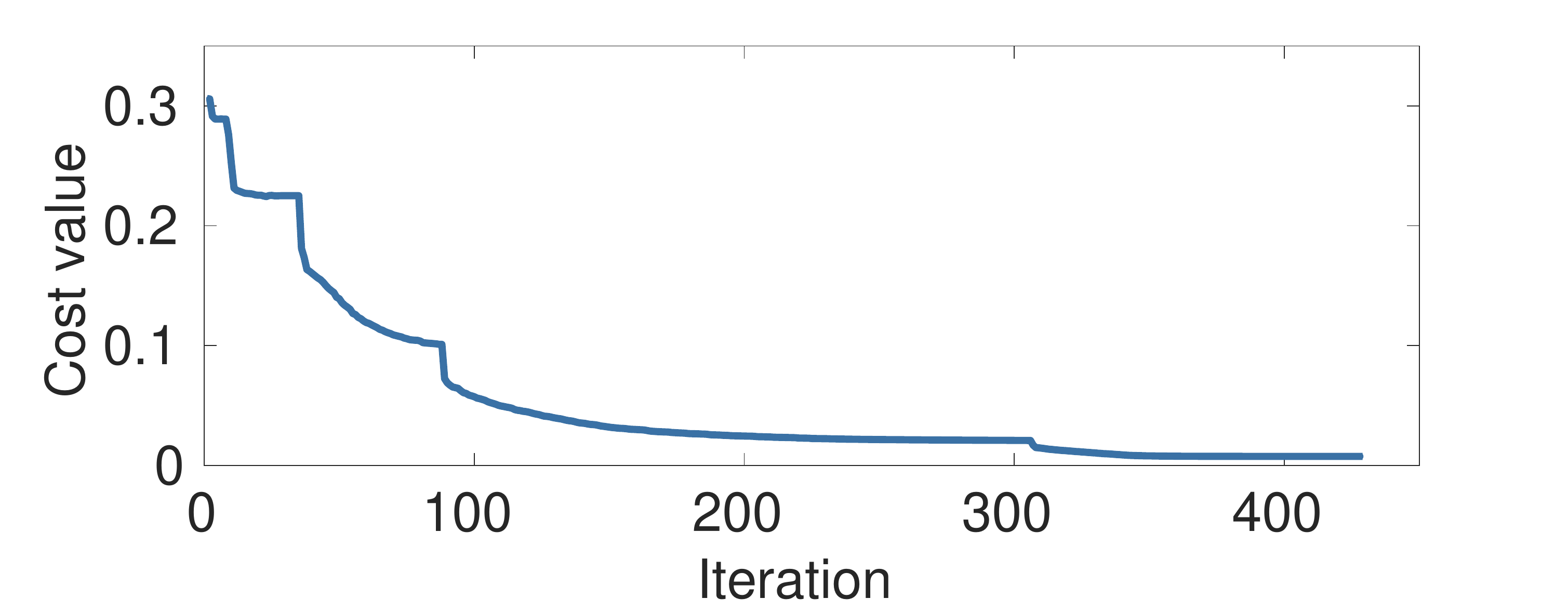}
        \caption{Figure shows a sample trace of the cost function \eqref{eq:9} over iterations, showing fast convergence.}
        \label{fig:opt_loss}
\end{figure}

In \eqref{eq:8}, $f(P \| Q)$ is a divergence/distance function between $P$ and $Q$. We use KL divergence because not only is it a natural choice, but we also find that it produces smooth distributions than when using the $\ell_2$ distance. Without a variance constraint, the solution of \eqref{eq:8} is simply a discrete delta function, which we would like to avoid. The variance constraint is equivalently expressed as a quadratic-term involving $p_i$'s, based on which we have: 
\begin{align}  \label{eq:9}
&\underset{P}{\text{min}} \, f(P \| Q) \nonumber \\
\text{s.t.}
 \, \sum_{i=1}^{n}p_i=1, \,\frac{1}{n}& \left( \sum_{i=1}^{n}i^2 p_i - \left(\sum_{i=1}^{n}i p_i\right)^2\right) \geq \xi, p_i \geq 0,
\end{align}
where, $\xi > 0$. In general the KL divergence is convex on the space of density functions, but in our case $P$ and $Q$ are related to each other; our objective function is not convex. We solve \eqref{eq:9} using {\em fmincon} in Matlab, which uses an interior-point algorithm with barrier functions. We note that the solution from fmincon may only be a locally optimal solution, yet we find the obtained solution is quite robust to large variation in initialization. We also note that using any of $KL(P\|Q)$, $KL(Q\|P)$ and $KL(P\|\frac{(P+Q)}{2})+KL(Q\|\frac{(P+Q)}{2})$ as objective in \eqref{eq:9} gives us the same result.
We use the following settings for {\em fmincon}: interior-point as algorithm, {\em Max Function Evaluations} $=4\times10^5$, {\em Max Iterations} $=10^5$ and $n=1024$. We use $\xi=0.75$ in our experiments because it provides the best FID score \cite{heusel2017gans}. Figure \ref{fig:opt_loss} shows the trace of the optimizer cost value for the above settings; we observe convergence to a local minimum in less than $500$ iterations. 

\begin{figure}[ht!]
     \centering
      \includegraphics[ width=\columnwidth]{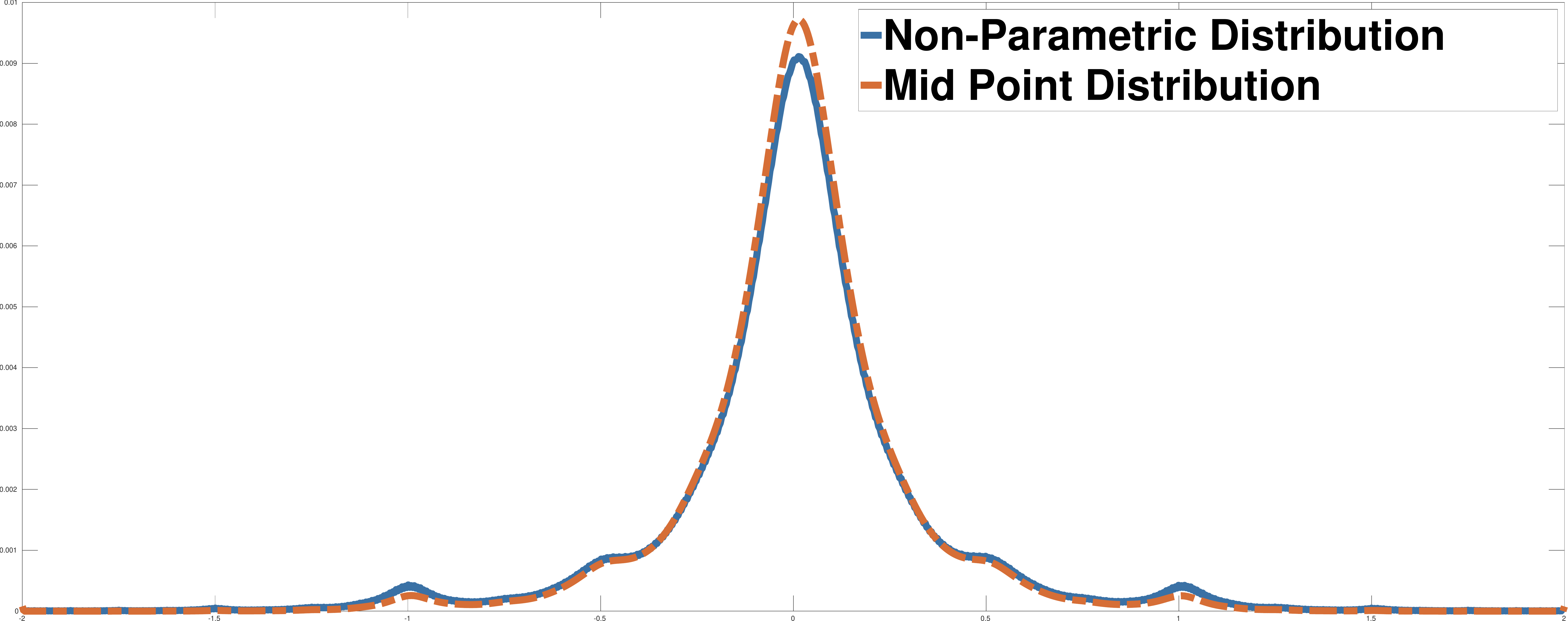}
        \caption{The distribution obtained by solving \eqref{eq:9} (shown in blue), and its mid-point distribution (red). While there is small variability in the solutions obtained, we find that no matter how we initialize the solver, all our obtained distributions share the three following traits: a large main-lobe, symmetry, and small side-lobes. Also, note the strong overlap between the distribution and the mid-point distribution. This is further quantified in table \ref{table:1} and compared with other distributions in figure \ref{fig:2}.}
        \label{fig:zoomed_in}
        \vspace{-0.2in}
\end{figure}

\paragraph{Remarks on the shape of the obtained distribution: } Here we make brief remarks on the shape of the obtained distribution. Firstly, we note that the exact shape of the obtained distribution varies slightly each time we run the solver. This is of course expected. However, we find that all obtained solutions seem to share the same general shape: they have a large main-lobe, appear to be symmetric, and have small but significant side-lobes. This is more clearly shown in Figure \ref{fig:zoomed_in}. We note that we did not impose any symmetry condition during optimization, yet these solutions emerged despite different initialization.  

\paragraph{Dependence on initialization: } We initialized our solver with a uniform density, delta functions centered at different locations, and truncated Gaussians with varying means and standard-deviations. For all these, the final solution still converges to a shape very similar to that shown in Figure \ref{fig:zoomed_in}. Further, all obtained solutions seem to perform equally well in the final evaluation of GAN output quality.

\paragraph{Role of side-lobes:} We are not aware of any simple parametric distribution that can describe the shape seen in Figure \ref{fig:zoomed_in}, except perhaps a Gaussian mixture model. However, the shape of the side-lobes is intricate, and not simple Gaussian-like. The existence of these side-lobes seems to allow us to strike a balance between the fast tail-decay of distributions like the Gamma, and the heavy tail of the Cauchy. It is almost as if the obtained shape fuses the best properties of the two classes of distributions, enabling us to generate good quality GAN output, all the while minimizing the divergence to the interpolated samples. 

\paragraph{Continuous samples from discretized density:} At first glance it may appear that since we discretize the domain $[0,1]$ while solving \eqref{eq:9}, that our prior is capable of generating only discrete samples. This is easily dealt with as follows. Once we generate a sample from the discretized density, what we get is really an index corresponding to the bin-center, but the bin itself has non-zero width given by how finely we partition $[0,1]$. From the corresponding bin, we simply generate a uniform random variable restricted to the width of the bin. This approach implicitly corresponds to sampling from a continuous density constructed by a zeroth-order interpolation over the obtained discrete one. One can be more sophisticated than this, but the sampling algorithm will no longer be as simple. We find that the approach described above is quite sufficient in practice. 

\paragraph{Quantification of mid-point mismatch:} Table \ref{table:1} shows the actual KL divergence between the prior and mid-point distribution. It is clear that for the normal and uniform distributions, the mid-point distribution is very divergent from the actual prior distribution; whereas the distribution obtained from solving \eqref{eq:9} has a much lower divergence from the mid-point. In \eqref{eq:9} we are only minimizing the distribution mismatch for the one-dimensional case. The idea is that if the 1-D distribution is similar to its mid-point distribution, then the divergence between the corresponding Euclidean norm distribution will be low even for higher dimensions.

\begin{figure*}[ht!]
     \centering
     \begin{subfigure}[]{0.32\textwidth}
         \centering
         \includegraphics[width=\textwidth]{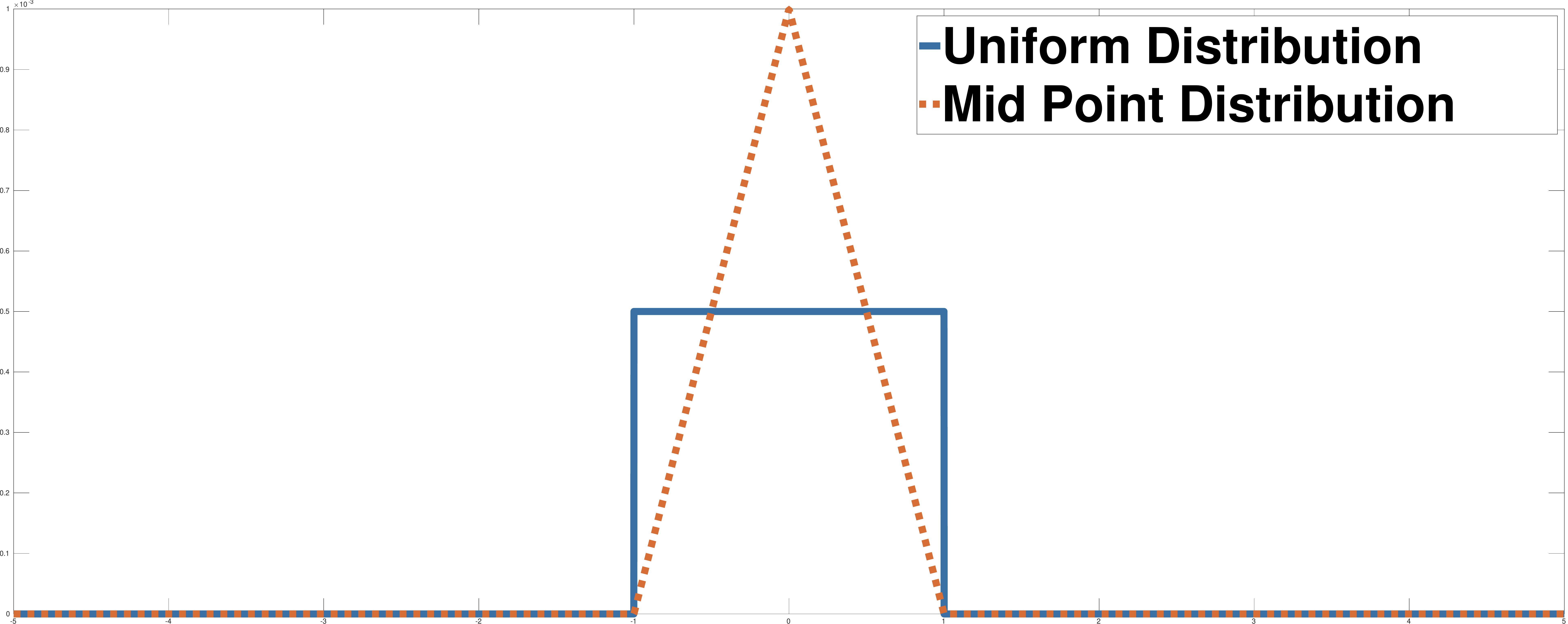}
         \caption{Uniform distribution}
     \end{subfigure}%
     \hfill
     \begin{subfigure}[]{0.32\textwidth}
         \centering
         \includegraphics[width=\textwidth]{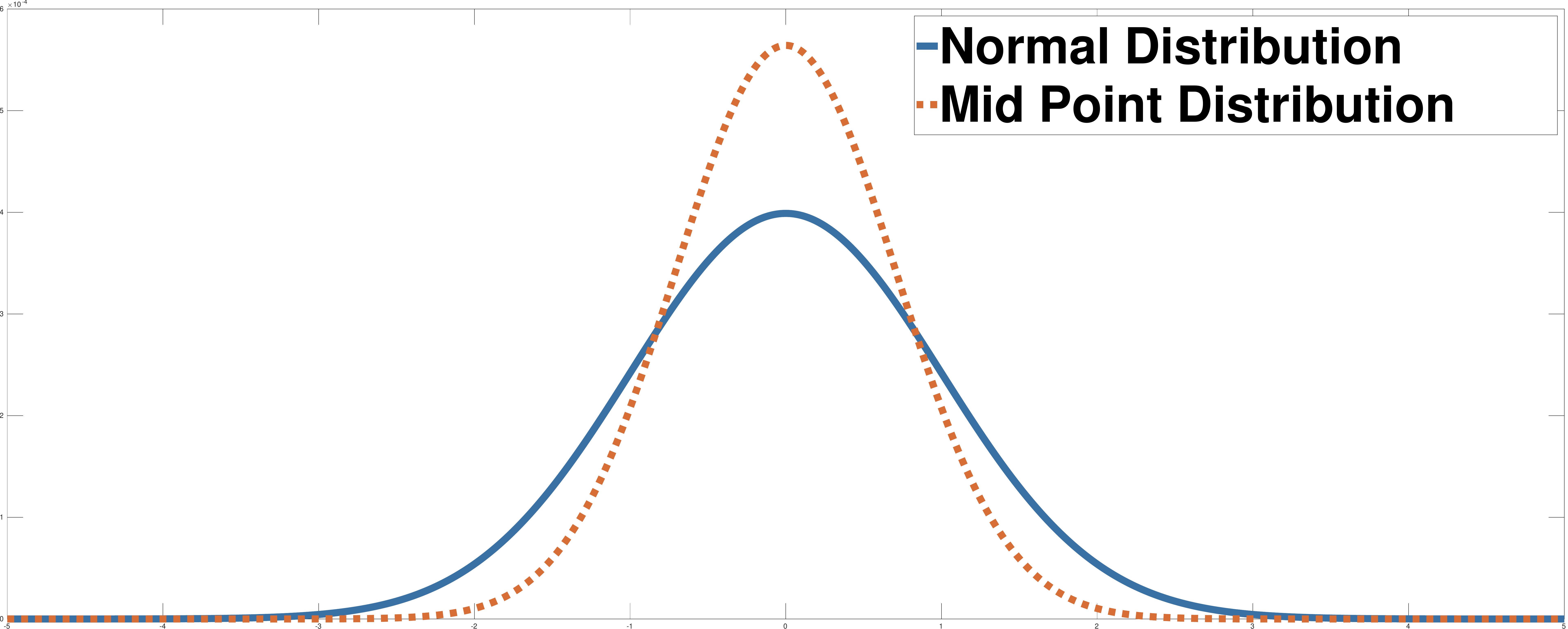}
         \caption{Normal distribution}
     
     \end{subfigure}%
     \hfill
     \begin{subfigure}[]{0.32\textwidth}
         \centering
         \includegraphics[width=\textwidth]{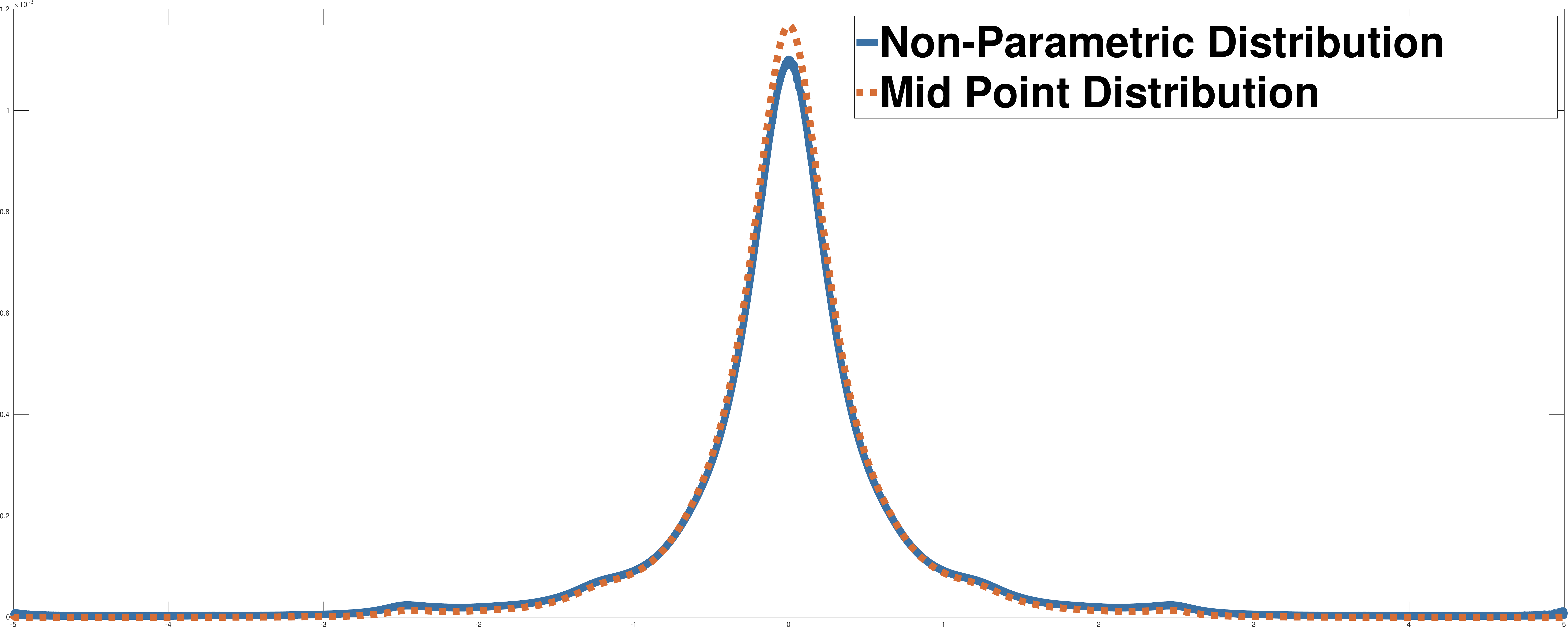}
         \caption{Obtained non-parametric distribution}
    
     \end{subfigure}
        \caption{The figures show various choices of priors (blue) and their corresponding mid-point distribution (red). Note that one can observe a large discrepancy between the prior and mid-point distributions, for typical choices such as the uniform and Normal. The prior we develop shows significantly less discrepancy. These discrepancies are also quantified via the KL-divergence in table \ref{table:1}.}
        \label{fig:1}
\end{figure*}

\begin{figure*}[ht!]
     \centering
     
     \hfill
     \begin{subfigure}[]{\textwidth}
         \centering
         \includegraphics[ width=\textwidth]{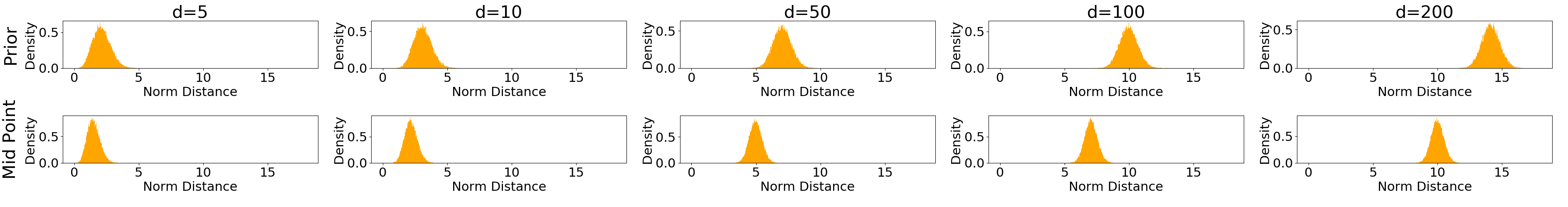}
         \caption{Normal distribution}
     
     \end{subfigure}%
     \hfill
     \begin{subfigure}[]{\textwidth}
         \centering
         \includegraphics[ width=\textwidth]{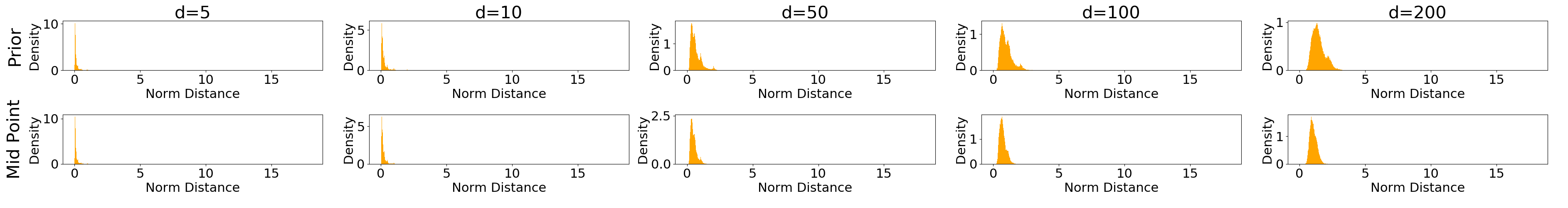}
         \caption{Obtained non-parametric distribution}
    
     \end{subfigure}
        \caption{Euclidean norm distribution for samples drawn from different priors and their corresponding mid-point Euclidean norm distribution for different dimensions $d$. 
        Note the mid-point norm distribution for the normal prior moves further away from the prior norm distribution as the dimension increases to $d=200$, whereas with our non-parametric prior, the mid-point norm distribution overlaps with the prior norm distribution even at $d=200$.}
        \label{fig:2}
        \vspace{-0.1in}
\end{figure*}

Figure \ref{fig:1} shows the mid-point distribution mismatch for different priors for the one dimensional case. 
Figure \ref{fig:2} shows the Euclidean norm distribution for prior and mid-points for different dimensions, computed from a set of $5 \times 10^4$ samples. For the mid-points, we sample two sets of $5\times 10^4$ points, and calculate the Euclidean norm of the corresponding mid-points. We see that for the normal distributions, at low dimensions ($d=5$ and $d=10$) the mid-point distribution overlaps well with the prior distribution. As the dimension increases ($d=50, 100, 200$), the two distributions start to diverge. For $d=100,200$ there is almost no overlap between the prior and mid-point distribution. We observe similar trend for the uniform distribution. 
 On the contrary, in our case the mid-point distribution and the prior distribution (of Euclidean norm) overlap well with each other even in higher dimensions. In Figure \ref{fig:2} we can notice that our non-parametric distribution does 
 bring the Euclidean norm distribution very close to the origin compared to the normal and uniform. 

\begin{table}[t]
\caption{KL-divergence between prior and mid-point distribution.}
\label{table:1}
\begin{small}

\begin{tabular}{lcccr}
\toprule
Distribution & KL divergence \\
\midrule
Uniform Distribution    &  0.3065\\
Normal Distribution & 0.1544\\
Proposed Non-Parametric Distribution    & {\bf 0.0075}\\

\bottomrule
\end{tabular}

\end{small}
\vspace{-0.1in}
\end{table}


\section{Experiments and Results}

\begin{figure*}[!htb]
     \centering
         \includegraphics[ width=0.99\textwidth]{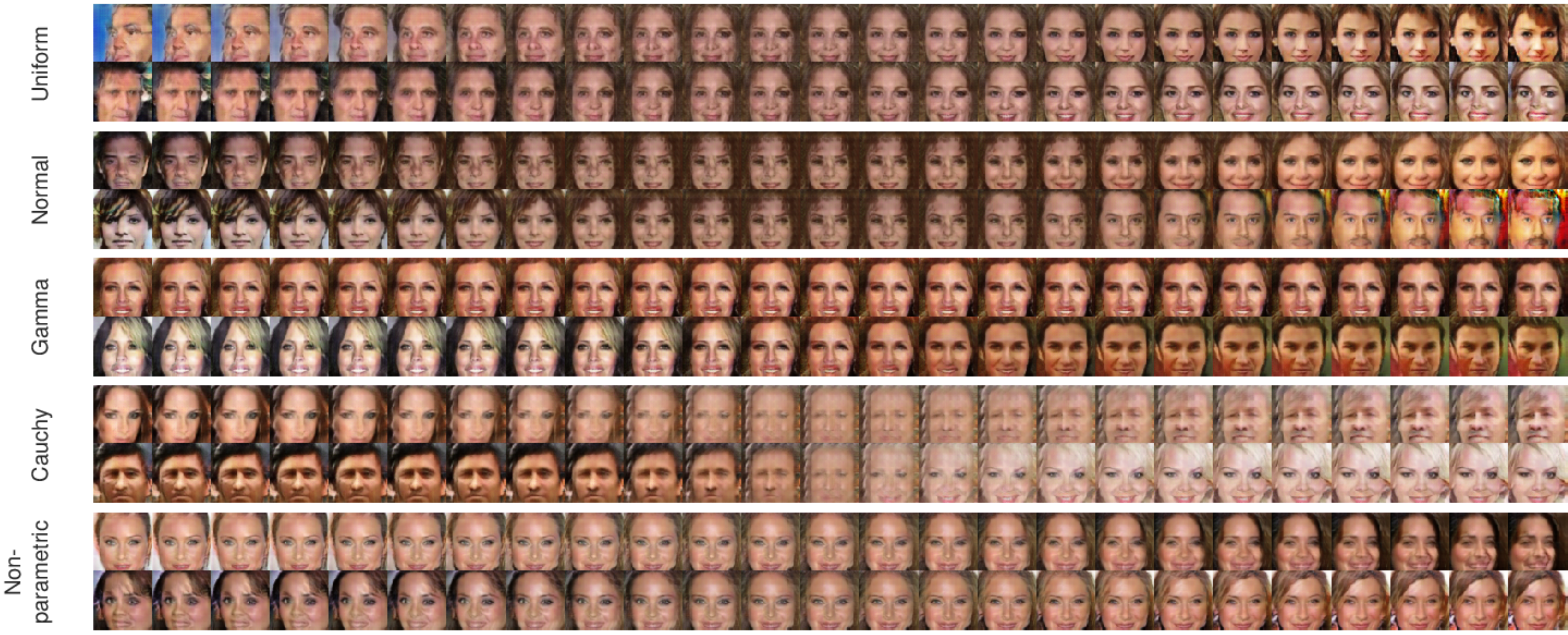}

        \caption{Interpolation (left to right) through the origin on CelebA dataset using different priors with $d=100$.  Note the degradation in image quality around the center of the panel (origin space) for many standard priors.}
        \label{fig:3}
\end{figure*}
     
    

\begin{figure*}[!htb]
     \centering
         \includegraphics[ width=0.96\textwidth]{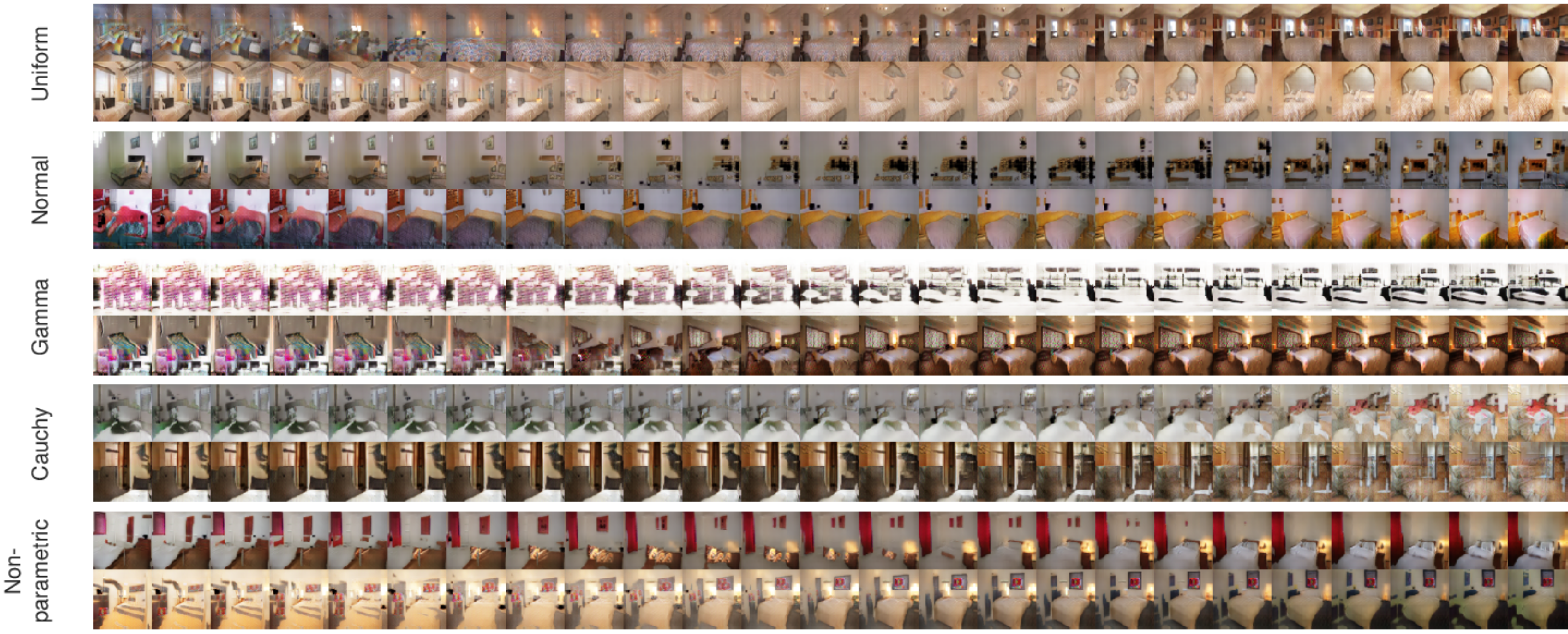}

        \caption{Interpolation (left to right) between two random points on LSUN bedroom dataset using different priors with $d=100$.}
        \label{fig:4}
\end{figure*}

\paragraph{Datasets, models, and baselines:} To validate the effectiveness of the proposed approach, we train the standard DCGAN model \cite{radford2015unsupervised} on four different datasets: a) CelebA dataset \cite{liu2015deep}, b) CIFAR10 \cite{krizhevsky2009learning}, c) LSUN Bedroom, and d) LSUN Kitchen \cite{yu2015lsun}). We train our model to the same number of epochs and all the hyper-parameters of the training are kept same for all the cases. We train each model three times and report the average scores. Details about the network architecture and the training method are provided in the supplemental material. We compare our proposed non-parametric prior distribution against standard ones like the normal, uniform and the priors designed to minimize the mid-point distribution mismatch like Gamma \cite{kilcher2017semantic}, and Cauchy \cite{lesniak2018latent}. For Gamma and Cauchy, we use the same parameters as suggested in the corresponding references.

 \paragraph{Qualitative tests: } In Figure \ref{fig:3}, we show the effect of interpolation through origin in high latent-space dimension ($d=100$) for different priors on CelebA datatset. Here, we interpolate between two random points such that the interpolation line passes through the origin. Similar to \cite{kilcher2017semantic}, we also observe that with standard priors like the normal, in high latent dimension, the GAN generates non-realistic images around the origin. Note the difference in quality near the images in the center of the panels (space around origin). With our non-parametric distribution obtained from the solution of \eqref{eq:9}, the GAN generates more realistic images around the origin even at higher dimensions. It was pointed out by Lesniak et. al. \cite{lesniak2018latent} that if a GAN is trained for more epochs, then it learns to fill the space around the origin even with the standard priors. We observe similar trend with the normal and uniform priors. However, with our non-parametric prior the GAN learns to fill the space around the origin very early in training compared with the standard priors. We also present qualitative comparisons on LSUN bedroom dataset in Figure \ref{fig:4}: comparing the results with the standard priors and the priors proposed in \cite{kilcher2017semantic} and \cite{lesniak2018latent}, highlighting the favorable performance of the proposed approach. While we note that it is difficult to perceptually appreciate whether we outperform the other priors, we show that we do obtain competitive visual quality with a conceptually general approach. In the supplemental material, we show additional qualitative results for LSUN Bedroom/Kitchen and CIFAR10 datasets. We note that the Cauchy distribution had difficulty converging on these datasets, exhibiting possible mode collapse and instability during GAN training. 

\paragraph{Quantitative evaluation: } For quantitative analysis, we use the Inception Score (IS) \cite{salimans2016improved} and the Frechet Inception Distance (FID) \cite{heusel2017gans} which are the standard metrics used to evaluate GAN performance. The inception score {\em correlates} with the visual quality of the generated image -- higher the better. However, recent studies suggest that the inception score does not compare the statistics of the generated dataset with the real-world data \cite{heusel2017gans,zhang2018self}, and thus is not always a reliable indicator of visual quality. This drawback of the IS is overcome by the FID score, which compares the statistics of the generated data with the real data with respect to features. The lower the FID score, the better. We will see in Tables \ref{table:4} to \ref{table:6}, that our non-parametric prior performs better in terms of FID score on both the prior and mid-point by at least $2$ points. In terms of IS we are the best in most of the cases, when we do not perform better, we come quite close to the best performing one. 

To get the IS and FID score we sample $5 \times 10^4$ points from the prior, and estimate the scores on the corresponding image samples. For mid-point, we sample two sets of $5 \times 10^4$ points from the prior, and an image is generated by the GAN, with the corresponding average points as inputs. Results are summarized in Tables \ref{table:4} to \ref{table:6} for different datasets. 
Table \ref{table:4} compares our non-parametric prior with other standard priors on the CelebA dataset, at latent space dimension $d=100$. The non-parametric prior outperforms all other priors on both the metrics. Our prior has better FID score by more than $6$ points on both the prior and the mid-point. As expected the IS and FID scores for the Cauchy is almost the same for the prior and the mid-point. With Gamma, we observe that the score on mid-points is slightly better than the prior since the gamma distribution is highly dense around the origin.
In Table \ref{table:ex}, we show the IS and FID for the prior and the mid-point at latent space dimension $d=200$. We note that our non-parametric distribution performs better in all the cases compared with all other priors. Scores for our non-parametric prior is almost similar to the scores in Table \ref{table:4}, which indicates its robustness toward the increase in latent space dimension. Cauchy prior sometimes leads to mode collapse during the training which is indicated by its poor FID scores. From Table \ref{table:4} and \ref{table:ex}, we also note that the IS and FID score become worse for the mid-point compared to the prior point as we increase the latent space dimension.
\begin{table}[t]
\caption{Comparison of IS and FID scores for different prior distributions on CelebA dataset with $d=100$}
\label{table:4}
\begin{small}
\begin{tabular}{|c|c|c|c|c|}
\hline
\multicolumn{1}{|c|}{Distribution} & \multicolumn{2}{|c|}{Inception Score}& \multicolumn{2}{|c|}{FID Score} \\
\hline
& \multicolumn{1}{|c|}{Prior} & \multicolumn{1}{|c|}{Mid-Point} & \multicolumn{1}{|c|}{Prior} & \multicolumn{1}{|c|}{Mid-Point} \\
 \hline
 Uniform & 1.843 & 1.369 &  24.055 & 40.371 \\
 \hline
 Normal & 1.805 & 1.371 & 26.173 & 42.136 \\
 \hline
 
 Gamma & 1.776 & 1.618 & 29.912 & 28.608 \\
 \hline
 Cauchy & 1.625 & 1.628 & 59.601 & 60.128 \\
 \hline
 
 {\bf Non-parametric} & {\bf 1.933} & {\bf 1.681} & {\bf 17.735} & {\bf 19.115} \\
 \hline
\end{tabular}
\end{small}
\vspace{-0.2in}
\end{table}
\begin{table}[t]
\caption{Comparison of IS and FID scores for different prior distributions on CelebA dataset with $d=200$}
\label{table:ex}
\begin{small}
\begin{tabular}{|c|c|c|c|c|}
\hline
\multicolumn{1}{|c|}{Distribution} & \multicolumn{2}{|c|}{Inception Score}& \multicolumn{2}{|c|}{FID Score} \\
\hline
& \multicolumn{1}{|c|}{Prior} & \multicolumn{1}{|c|}{Mid-Point} & \multicolumn{1}{|c|}{Prior} & \multicolumn{1}{|c|}{Mid-Point} \\
 \hline
 Uniform & 1.908 & 1.407 & 25.586 & 44.837 \\
 \hline
 Normal & 1.857 &  1.434 & 25.035 & 43.596 \\
 \hline
 
 Gamma & 1.738  & 1.608 & 33.816 & 32.241 \\
 \hline
 Cauchy & 1.734 & {\bf 1.743} & 86.286 & 86.278 \\
 \hline
 
 {\bf Non-parametric} & {\bf 1.973 } &  1.636  &  {\bf 14.953}  & {\bf 19.322}  \\
 \hline
\end{tabular}
\end{small}
\vspace{-0.1in}
\end{table}
\begin{table}[H]
\vspace{-0.3in}
\caption{Comparison of IS and FID scores for different prior distributions on CIFAR10 dataset with $d=100$}
\label{table:7}
\begin{small}
\begin{tabular}{|c|c|c|c|c|}
\hline
\multicolumn{1}{|c|}{Distribution} & \multicolumn{2}{|c|}{Inception Score}& \multicolumn{2}{|c|}{FID Score} \\
\hline
& \multicolumn{1}{|c|}{Prior} & \multicolumn{1}{|c|}{Mid-Point} & \multicolumn{1}{|c|}{Prior} & \multicolumn{1}{|c|}{Mid-Point} \\
 \hline
 Uniform & 6.411 & 5.204 &  43.501 & 76.913 \\
 \hline
 Normal & 6.836 & 5.656 & 39.235 & 65.525 \\
 \hline
 
 Gamma & 6.449 & 6.798 & 48.334 & 39.262 \\
 \hline
 Cauchy & 2.972 & 2.964 & 180.37 & 180.40\\
 \hline
 
 {\bf Non-parametric} & {\bf 6.871} & {\bf 6.809} & {\bf 34.803} & {\bf 37.112} \\
 \hline
 
\end{tabular}
\end{small}
\vspace{-0.1in}
\end{table}
\begin{table}[t]
\caption{Comparison of IS and FID scores with different prior distributions on LSUN Bedroom dataset with $d=100$}
\label{table:5}
\begin{small}
\begin{tabular}{|c|c|c|c|c|}
\hline
\multicolumn{1}{|c|}{Distribution} & \multicolumn{2}{|c|}{Inception Score}& \multicolumn{2}{|c|}{FID Score} \\
\hline
& \multicolumn{1}{|c|}{Prior} & \multicolumn{1}{|c|}{Mid-Point} & \multicolumn{1}{|c|}{Prior} & \multicolumn{1}{|c|}{Mid-Point} \\
 \hline
 Uniform & 2.969 & 2.649 & 42.998 & 76.412 \\
 \hline
 Normal & 2.812 & 2.591 & 64.682 & 108.49 \\
 \hline
 
 Gamma & 2.930 & 2.808 & 162.44 & 161.37 \\
 \hline
 Cauchy & {\bf 3.148} & {\bf 3.149} &  97.057 & 97.109 \\
 \hline
 
 {\bf Non-parametric} & 3.028 & 2.769 & {\bf 27.857} & {\bf  31.472} \\
 \hline
\end{tabular}
\end{small}
\vspace{-0.2in}
\end{table}
\begin{table}[t]
\vspace{-0.2in}
\caption{Comparison of IS and FID scores with different prior distributions on LSUN Kitchen dataset with $d=100$}
\label{table:6}
\begin{small}
\begin{tabular}{|c|c|c|c|c|}
\hline
\multicolumn{1}{|c|}{Distribution} & \multicolumn{2}{|c|}{Inception Score}& \multicolumn{2}{|c|}{FID Score} \\
\hline
& \multicolumn{1}{|c|}{Prior} & \multicolumn{1}{|c|}{Mid-Point} & \multicolumn{1}{|c|}{Prior} & \multicolumn{1}{|c|}{Mid-Point} \\
 \hline
 Uniform & 2.656 & 2.549 & 40.041 & 51.119 \\
 \hline
 Normal & 2.844 & 2.867  & 39.909 & 53.448 \\
 \hline
 
 Gamma & 2.183 & 2.147 & 181.81 & 187.00 \\
 \hline
 Cauchy & 1.182 & 1.179 & 242.27 & 242.87 \\
 \hline
 
 {\bf Non-parametric} & {\bf 3.109} & {\bf 3.031}  & {\bf 33.194 } & { \bf 35.074 } \\
 \hline
\end{tabular}
\end{small}
\vspace{-0.1in}
\end{table}
 Table \ref{table:7} shows the IS and FID scores for CIFAR10. With our non-parametric prior, the GAN performs better than other priors on both the prior and mid-point by at least $2$ points on FID score. Similar to CelebA dataset, we observe the training with Cauchy prior is highly unstable. In Table~\ref{table:5} and~\ref{table:6} we compare our non-parametric prior with other priors on the LSUN bedroom and LSUN kitchen datasets. We observe that our non-parametric prior outperform other priors on the FID score by at least $6$ points. We observe that the Gamma and Cauchy priors perform worse on both the prior and mid-point in terms of FID score. These priors often lead to mode collapse during the training. Note that the LSUN dataset has a larger  variation in images, and also a larger training-set than the CelebA dataset. The non-parametric prior performs best in both cases as measured by the FID  on both prior and mid-point, showing its benefits on large datasets with large variation. 
{\bf A few salient observations} from the results are:
\begin{compactitem}
\item The quantitative results on different datasets show that standard priors like the normal and uniform perform better on the prior point but worse on the mid-point. 
\item The priors proposed to minimize the mid-point distribution mismatch in \cite{kilcher2017semantic} and \cite{lesniak2018latent} achieve better results on the mid-point but perform worse on the prior point. 
\item Gamma and Cauchy do not perform consistently across datasets. In some cases they are the best, but when they are not, their performance can be far from the best. 
\item The non-parametric distribution is far more consistent, and is either the best, or pretty close to the best performing one, on all four datasets.
\end{compactitem}

\section{Conclusions}
In this paper, we propose a generalized approach to solve distribution mismatch for interpolation in GANs.  We show the qualitative and quantitative effectiveness of our approach over the standard priors. We note that often times, our proposed method is in fact the best one, and in cases when it is not, it comes quite close to the best performing technique. Our goal is not necessarily outperform all other GANs, but to suggest the use of non-trivial priors, which might improve image quality without any additional training-data or architectural complexity. Additionally, it would be an interesting avenue of future work to extend this approach to extrapolation problems, or impose other interesting statistical or physically-motivated constraints over latent spaces.

\section*{Acknowledgements}
This work was supported by a Intel HVMRC grant. PT was supported by ARO grant number W911NF-17-1-0293. SJ was jointly supported from both the Herberger Research Initiative in the Herberger Institute for Design and the Arts (HIDA) and the Fulton Schools of Engineering (FSE) at Arizona State University.

\bibliographystyle{icml2019}
\bibliography{main}

\end{document}